\documentclass[a4paper]{article}
\usepackage{INTERSPEECH2015,amssymb,amsmath,epsfig,footnote,graphicx,cite}

\usepackage[tight,footnotesize]{subfigure}

\title{Sequence-to-Sequence Neural Net Models for \\Grapheme-to-Phoneme Conversion}

\makeatletter
\def\name#1{\gdef\@name{#1\\}}
\makeatother
\makeatother \name{{\em Kaisheng Yao, Geoffrey Zweig}}

\address{Microsoft Research\\
{\small \tt \{kaisheny, gzweig\}@microsoft.com}}

\ninept

\begin{document}

\maketitle

\begin{abstract}

Sequence-to-sequence translation methods based on generation
with a side-conditioned language model have recently
shown promising results in 
several tasks. In machine translation, models conditioned 
on source side
words have been used to produce target-language text, and 
in image captioning, models
conditioned images have been used to generate
caption text. Past work with this
approach has focused on large vocabulary tasks, and measured quality in terms
of BLEU. In this paper, we explore the applicability of such models to 
the qualitatively different grapheme-to-phoneme task. Here, the input and
output side vocabularies are small, plain n-gram models do well,
and credit is only given when the
output is exactly correct. We find that the simple side-conditioned generation
approach is able to rival the state-of-the-art, and we are 
able to significantly advance the stat-of-the-art with bi-directional long short-term memory (LSTM) neural networks that use the same alignment information that is used in conventional
approaches.
\end{abstract}

\section{Introduction}

In recent work on sequence to sequence translation, it has been shown 
that side-conditioned neural networks can be effectively used for both 
machine translation \cite{son2012continuous,auli2013joint,KalchbrennerEMNLP2013,DevlinNNJSMT,SutskeverSequenceToSequence2014,SundermeyerTranslation} 
and image captioning 
\cite{fang2014captions,karpathy2014deep,vinyals2014show,donahue2014long}. 
The use of a side-conditioned language model \cite{Mikolov+Zweig2012} is 
attractive for its simplicity, and apparent performance, and 
these successes
complement other recent work in which neural
networks have advanced the state-of-the-art, for example in
language modeling~\cite{BengioNNLM2003,MikolovASRU2011}, language understanding~\cite{KY+RNN+2013}, and parsing~\cite{VinyalsGrammar2014}. 

In these previously studied tasks, the input vocabulary size is large, and the statistics for many words must be sparsely estimated. 
To alleviate this problem, neural network based approaches use 
continuous-space representations of words, in which words that occur in similar contexts tend to be close to each other in representational space. 
Therefore, data that benefits one word in a particular context 
causes the model to generalize to similar words in similar contexts. The benefits of using neural networks, in particular, both
simple recurrent neural networks~\cite{MikolovThesis} 
and long short-term memory (LSTM) neural networks~\cite{Hochreiter97,Gers99learningto,AlexLSTM}, to deal with sparse statistics are very apparent. 

However, to our best knowledge, the top performing methods for the
grapheme-to-phoneme (G2P) task have been
based on the use of
Kneser-Ney n-gram models~\cite{BisaniGraphone2008}.
Because of the 
relatively small cardinality of letters and phones, n-gram statistics,
 even with long context windows, can be reliably trained. 
On G2P tasks, maximum entropy models~\cite{Chen03G2P}
 also perform well. 
The G2P task is distinguished in another important way: whereas the
machine translation and image captioning tasks are scored
with the relatively forgiving BLEU metric, in the G2P task, a phonetic
sequence must be exactly correct in order to get credit when scored.

In this paper, we study the open question of whether side-conditioned generation
approaches are competitive on the grapheme-to-phoneme task.
We find that LSTM approach
proposed by \cite{SutskeverSequenceToSequence2014} performs well and is very close to the state-of-the-art. While the
side-conditioned LSTM approach does not require any alignment information,
the state-of-the-art ``graphone'' method of ~\cite{BisaniGraphone2008}
is based on the use of 
alignments. We find that when we allow the neural network approaches to also
use alignment information, we significantly advance the  state-of-the-art.

The remainder of the paper is structured as follows. We review previous 
methods in Sec.~\ref{sec:background}. We then present side-conditioned
generation models in Sec.~\ref{sec:scgen}, and
models that leverage alignment information in Sec.~\ref{sec:fbf}. 
We present experimental results in Sec.~\ref{sec:experiments} and
provide a further comparison with past work in Sec.~\ref{sec:related}. We
conclude in Sec.~\ref{sec:conclusions}.

\section{Background}
\label{sec:background}
This section summarizes the state-of-the-art solution for G2P conversion. The G2P conversion can be viewed as translating an input sequence of graphemes
(letters)
 to an output sequence of phonemes. Often, the grapheme and phoneme sequences have been aligned to form joint grapheme-phoneme units. 
In these alignments, a grapheme may correspond to a null phoneme with no pronunciation, a single phoneme,
 or a compound phoneme. The compound phoneme is a concatenation of two phonemes. An example is given in Table~\ref{tab:ltsexample}.
\begin{table}[h]
\begin{tabular}{c|cccccc}
$Letters$ & T &A &N &G &L    &E \\
$Phonemes$ & T &AE&NG&G &AH:L &null
\end{tabular}
\caption{An example of an alignment of letters to phonemes. The letter L aligns to a compound phoneme, and the letter E to a null phoneme that is not
pronounced. \label{tab:ltsexample}}
\end{table}

Given a grapheme sequence $L = l_1,\cdots,l_T$, a corresponding  phoneme sequence $P = p_1, \cdots, p_T$, and an alignment $A$, the posterior probability $p(P|L,A)$ is approximated as: 
\begin{eqnarray}
p(P|A,L) & \approx & \prod_{t=1}^T p(p_t|p_{t-k}^{t-1}, l_{t-k}^{t+k}) \label{eqn:postprob}
\end{eqnarray}
\noindent where $k$ is the size of a context window, and $t$ indexes the 
positions in the alignment. 

Following \cite{BergerMEmodel1996,Chen03G2P}, Eq. (\ref{eqn:postprob}) can be estimated using an exponential (or \textit{maximum entropy}) model in the form of 
\begin{equation}
p(p_t|x=(p_{t-k}^{t-1}, l_{t-k}^{t+k})) = \frac{\exp (\sum_i \lambda_i f_i(x, p_t))}{\sum_q \exp (\sum_i \lambda_i f_i(x,q))}
\end{equation}
\noindent where features $f_i(\cdot)$ are usually 0 or 1 indicating the identities of phones and letters in specific contexts. 

Joint modeling has been proposed for grapheme-to-phoneme conversion~\cite{GalescuG2P,Chen03G2P,BisaniGraphone2008}. In these models, one has a vocabulary of grapheme and phoneme pairs, which are called graphones. The probability of a graphone sequence is 
\begin{equation}
p(C=c_1 \cdots c_T)  = \prod_{t=1}^T p(c_t|c_1 \cdots c_{t-1}), \label{eqn:chunk}
\end{equation}
where each $c$ is a graphone unit.
The conditional probability $p(c_t|c_1 \cdots c_{t-1})$ is estimated using an n-gram language model.

To date, these models have produced
 the best performance on common benchmark datasets, 
and are used for comparison with the architectures in the following sections.

\section{Side-conditioned Generation Models}
\label{sec:scgen}

In this section, we explore the use of side-conditioned language
models for generation. This approach is appealing for its simplicity, and
especially because no explicit  alignment information is needed.

\subsection{Encoder-decoder LSTM}
\label{sec:encoderdecoder}
\begin{figure}[t]
\begin{center}
  \centerline{\includegraphics[width=8.0cm]{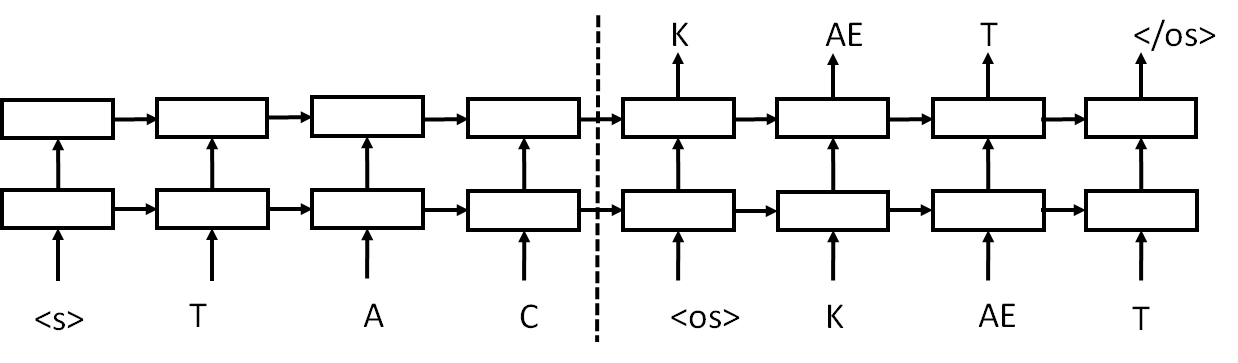}}
\end{center}
\vspace{-0.3cm}
\caption{An encoder-decoder LSTM with two layers. The encoder LSTM, to the left of the dotted line, reads a time-reversed sequence ``$\langle s \rangle$ T A C'' and produces the last hidden layer activation to initialize the decoder LSTM. The decoder LSTM, to the right of the dotted line, reads ``$\langle os \rangle$ K AE T" as the past phoneme prediction sequence and uses "K AE T $\langle /os \rangle$'' as the output sequence to generate. Notice that the input sequence for encoder LSTM is time reversed, as in \cite{SutskeverSequenceToSequence2014}. $\langle s \rangle$ denotes letter-side sentence beginning. $\langle os \rangle$ and $\langle /os \rangle$ are the output-side sentence begin and end symbols. 
\label{fig:encoderdecoderlstm}}
\end{figure}

In the context of general sequence to sequence learning,
the concept of encoder and decoder networks has recently been proposed 
~\cite{KalchbrennerEMNLP2013,PhraseLearning,AlexLSTM,NeuralTranslation,SutskeverSequenceToSequence2014}. The main idea is mapping the entire input sequence to a vector, and then using a recurrent neural network (RNN) to generate the output sequence conditioned on 
the encoding vector. Our implementation follows the method in~\cite{SutskeverSequenceToSequence2014}, which we denote as encoder-decoder LSTM. Figure~\ref{fig:encoderdecoderlstm} depicts a model of this method. 
As in \cite{SutskeverSequenceToSequence2014}, we use an LSTM~\cite{AlexLSTM} as the basic recurrent network unit because it has shown better performance than simple RNNs on language understanding~\cite{YaoLSTMLU} and acoustic modeling~\cite{GoogleLSTM} tasks. 

In this method, there are two sets of LSTMs: one is an encoder that reads the source-side input sequence and the other is a decoder that functions as a language model
and generates the output. 
The encoder is used to represent the entire input sequence in the last-time hidden layer activities. 
These activities are used as the initial activities of the decoder
network.
The decoder is a language model that uses past phoneme sequence $\phi_1^{t-1}$ to predict the next phoneme $\phi_t$, with its hidden state initialized as described. 
It stops predicting after outputting $\langle /os \rangle$, the output-side end-of-sentence symbol. 
Note that in our models, we use $\langle s \rangle$ and $\langle /s \rangle$ as input-side 
begin-of-sentence and 
end-of-sentence tokens, and $\langle os \rangle$ and $\langle /os \rangle$ for corresponding output symbols.

To train these encoder and decoder networks, we used back-propagation through time (BPTT)~\cite{BPTTRobinson,BPTT}, with the error signal originating in the
decoder network. 

We use a beam search decoder to generate phoneme sequence during the decoding phase. 
The hypothesis sequence with the highest posterior probability is selected as the decoding result.

\section{Alignment Based Models}
\label{sec:fbf}

In this section, we relax the earlier constraint that the model translates
directly from the source-side letters to the target-side phonemes without
the benefit of an explicit alignment.

\subsection{Uni-directional LSTM}
\label{sec:unidirectional}

\begin{figure}[t]
\begin{center}
  \centerline{\includegraphics[width=8.0cm]{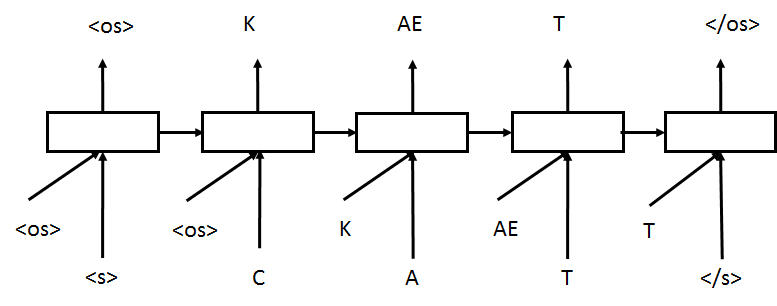}}
\end{center}
\vspace{-0.3cm}
\caption{The uni-directional LSTM reads letter sequence ``$\langle s \rangle$ C A T $\langle /s \rangle$'' and past phoneme prediction ``$\langle os \rangle$ $\langle os \rangle$ K AE T''. It outputs phoneme sequence ``$\langle os \rangle$ K AE T $\langle /os \rangle$''. Note that there are separate output-side begin and end-of-sentence symbols, prefixed by "o".}
\label{fig:unidirection}
\end{figure}

A model of the uni-directional LSTM is in Figure~\ref{fig:unidirection}.
Given a pair of source-side input and target-side output sequences and an 
alignment $A$, 
the posterior probability of output sequence given the input sequence is
\begin{equation}
p(\phi_1^T |A, l_1^T) = \prod_{t=1}^T p(\phi_t | \phi_1^{t-1}, l_1^t)\label{eqn:jointconditional}
\end{equation}
\noindent where the current phoneme prediction $\phi_t$ depends both on its past prediction $\phi_{t-1}$ and the input letter sequence $l_t$. Because of 
the recurrence in the LSTM, prediction of the current phoneme depends on the phoneme predictions and letter sequence from the sentence beginning. 
Decoding uses the same beam search decoder described in Sec.~\ref{sec:scgen}.

\subsection{Bi-directional LSTM}
\label{sec:sbjcrnn}
The bi-directional recurrent neural network was proposed in \cite{BRNN}. 
In this architecture, one RNN processes the input from left-to-right, while
another processes it right-to-left. The outputs of the two sub-networks
are then combined, for example being fed into a third RNN.
The idea has been used for speech recognition~\cite{BRNN} and more recently for language understanding~\cite{MesnilInterSub13}. Bi-directional LSTMs have been applied to speech recognition~\cite{AlexLSTM} and machine translation~\cite{SundermeyerTranslation}. 

In the bi-directional model, the phoneme prediction depends on the whole source-side letter sequence as follows
\begin{equation}
p(\phi_1^T | A, l_1^T) = \prod_{t=1}^T p(\phi_t | \phi_1^{t-1} l_1^T) \label{eqn:sbjc}
\end{equation}

\begin{figure}[t]
\begin{center}
  \centerline{\includegraphics[width=8.0cm]{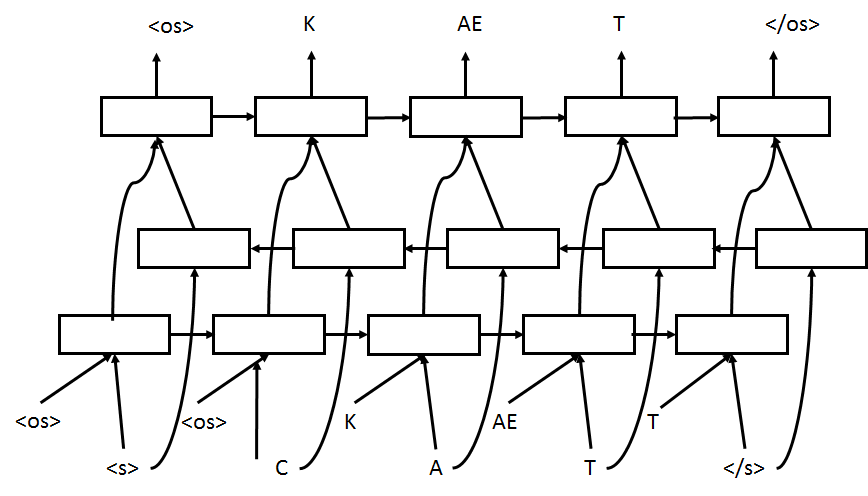}}
\end{center}
\vspace{-0.3cm}
\caption{The bi-directional LSTM reads letter sequence ``$\langle s \rangle$ C A T $\langle /s \rangle$'' for the forward directional LSTM, the time-reversed sequence ``$\langle /s \rangle$ T A C $\langle s \rangle$'' for the backward directional LSTM, and past phoneme prediction ``$\langle os \rangle$ $\langle os \rangle$ K AE T''. It outputs phoneme sequence ``$\langle os \rangle$ K AE T $\langle /os \rangle$''. }
\label{fig:bidirection}
\end{figure}

Figure~\ref{fig:bidirection} illustrates this model. Focusing on the third set of inputs, for example, 
letter $l_t = A$ is projected to a hidden layer, together with the past phoneme prediction $\phi_{t-1} = K$. The letter $l_t = A$ is also projected to a hidden layer in the network that runs in the backward direction. 
The hidden layer activation from the forward and backward networks
is then used as the input to a final network running in the forward direction. 
The output of the topmost recurrent layer is used to predict the current phoneme $\phi_t = AE$. 

We found that performance is better when feeding the past phoneme prediction to the bottom LSTM layer, instead of other layers such as the softmax layer. However, this architecture can be further extended, e.g., by feeding the past phoneme predictions to \textit{both} the top and bottom layers, which we may investigate in future work.

In the figure, we draw one layer of bi-directional LSTMs. 
In Section \ref{sec:experiments}, we also report results for deeper networks,
in which the forward and backward layers are duplicated several times; each
layer in the stack takes the concatenated outputs of the forward-backward 
networks below as its input.

Note that the backward direction LSTM is independent of the past phoneme predictions. Therefore, during decoding, we first pre-compute its activities. We then treat the output from the backward direction LSTM as additional input to the top-layer LSTM that also has input from the lower layer forward direction LSTM. The same beam search decoder described before can then be used.

\section{Experiments}
\label{sec:experiments}
\subsection{Datasets}
Our experiments were conducted on the three US English datasets\footnote{We thank Stanley F. Chen who kindly shared the data set partition he used in \cite{Chen03G2P}.}: the CMUDict, NetTalk, and Pronlex datasets that have been evaluated in \cite{Chen03G2P,BisaniGraphone2008}. We report phoneme error rate (PER) and word error rate (WER)~\footnote{We observed a strong correlation of BLEU and WER scores on these tasks. Therefore we didn't report BLEU scores in this paper. }. In the phoneme error rate computation, following~\cite{Chen03G2P,BisaniGraphone2008}, in the case of multiple reference pronunciations, the variant with the smallest edit distance is used. Similarly, if there are multiple reference pronunciations for a word, a word error occurs only if the predicted pronunciation doesn't match any of the references. 

The CMUDict contains 107877 training words, 5401 validation words, and 12753 words for
testing. 
The Pronlex data contains 83182 words for training, 1000 words for validation, and 4800 words for testing.
The NetTalk data contains 14985 words for training and 5002 words for testing, and does not have a validation set.

\subsection{Training details}

For the CMUDict and Pronlex experiments, all meta-parameters were set via experimentation with the 
validation set. For the NetTalk experiments, we used the same model structures 
as with the Pronlex experiments.

To generate the alignments used for training the alignment-based methods of Sec.~\ref{sec:fbf}, we used the alignment package of \cite{jiampojamarn2007}.
We used BPTT to train the LSTMs. We used sentence level minibatches without truncation. 
To speed-up training, we used data parallelism with 100 sentences per 
minibatch,
except for the CMUDict data, where one 
sentence per minibatch gave the best performance
on the development data.
For the alignment-based methods, we sorted sentences according to their lengths, and each minibatch had sentences with the same length. 
For encoder-decoder LSTMs, we didn't sort sentences in the same lengths as done in the alignment-based methods, and instead, followed \cite{SutskeverSequenceToSequence2014}. 

For the encoder-decoder LSTM in Sec.~\ref{sec:scgen}, we used 500 dimensional
projection and hidden layers. 
When increasing the depth of
the encoder-decoder LSTMs, we increased the depth of both encoder and decoder
networks.
For the bi-directional LSTMs, we used a 50 dimensional projection layer and 300 dimensional hidden layer.
For the uni-directional LSTM experiments on CMUDict, we used a 400 dimensional projection layer, 400 dimensional hidden layer, and the above described data parallelism. 

For both encoder-decoder LSTMs and the alignment-based methods, we randomly permuted the order of the training sentences in each epoch. 
We found that the encoder-decoder LSTM needed to start from a small learning rate, approximately 0.007 per sample. For bi-directional LSTMs, we used initial learning rates of  0.1 or 0.2. For the uni-directional LSTM, the initial learning rate was 0.05.
The learning rate was controlled by monitoring the improvement of
cross-entropy scores on validation sets. If there was no improvement of the cross-entropy score, we halved the learning rate.
NetTalk dataset doesn't have a validation set. Therefore, on NetTalk, we first ran
10 iterations with a fixed per-sample learning rate of 0.1, reduced the
learning rate by half for 2 more iterations, and finally used 0.01
for 70 iterations. 

The models of Secs.~\ref{sec:scgen} and \ref{sec:fbf} require using a beam
search decoder. Based on validation results, we report results with beam width of 1.0 in likelihood. 
We did not observe an improvement with larger beams.
Unless otherwise noted, we used a window of 3 letters in the models.
We plan to release our training recipes to public through computation network toolkit (CNTK)~\cite{CNTK}.

\subsection{Results}
\label{sec:cmuresult}
We first report results for all our models on the CMUDict dataset 
\cite{Chen03G2P}. 
The first two lines of Table \ref{tab:cmu} show results for the encoder-decoder
models. While the error rates are reasonable, 
the best previously reported results of 24.53\% WER 
\cite{BisaniGraphone2008} are somewhat better. It is possible that 
combining multiple systems as in \cite{SutskeverSequenceToSequence2014} would achieve the same 
result, we have chosen not to engage in system combination.

The effect of using alignment based models is shown at the bottom of 
Table \ref{tab:cmu}. Here, the bi-directional models produce an unambiguous improvement over the earlier 
models, and by training a three-layer bi-directional LSTM, we are able to significantly 
exceed the previous state-of-the-art. 

We noticed that the uni-directional LSTM with default window size had the highest WER, perhaps because one does not observe the entire input sequence as is 
the case with both the encoder-decoder and bi-directional LSTMs. To validate this claim, we increased the window size to 6 to include the current and five future letters as its source-side input. Because the average number of letters is 7.5 on CMUDict dataset, the uni-directional model in many cases thus
sees the entire letter sequences. With a window size of 6 and additional information from the alignments, the uni-directional model was able to perform better than the encoder-decoder LSTM. 

\begin{table}[t]
\centering
\begin{tabular}{|l|c|c|}
\hline
Method & PER (\%) & WER (\%) \\
\hline
\hline
encoder-decoder LSTM & 7.53 & 29.21 \\
encoder-decoder LSTM (2 layers) & 7.63 & 28.61  \\
\hline
\hline
uni-directional LSTM & 8.22 & 32.64 \\
uni-directional LSTM (window size 6) & 6.58& 28.56 \\
\hline
bi-directional LSTM & 5.98 & 25.72   \\
bi-directional LSTM (2 layers) & 5.84 & 25.02  \\
bi-directional LSTM (3 layers) & 5.45 & 23.55  \\
\hline
\hline
\end{tabular}
\caption{Results on the CMUDict dataset. }
\label{tab:cmu}
\end{table}

\subsection{Comparison with past results}
\label{exp:mainresults}

We now present additional results for the NetTalk and Pronlex datasets,
and compare with the best previous results.
The method of \cite{BisaniGraphone2008} uses 9-gram graphone models, and
\cite{Chen03G2P} uses 8-gram maximum entropy model. 

Changes in WER of 
0.77, 1.30, and 1.27 for CMUDict, NetTalk and
Pronlex datasets respectively are significant at the 95\% confidence level. 
For PER, the corresponding values are  0.15, 0.29, and 0.28.
On both the CMUDict and NetTalk datasets, the bi-directional LSTM outperforms the previous results at the 95\% significance level. 

\begin{table}[t]
\centering
\begin{tabular}{|l|c|c|c|}
\hline
Data & Method & PER (\%) & WER (\%) \\
\hline \hline
CMUDict & past results~\cite{BisaniGraphone2008} & 5.88 & 24.53  \\
& bi-directional LSTM & 5.45 & 23.55 \\
\hline \hline
NetTalk & past results~\cite{BisaniGraphone2008} & 8.26 & 33.67 \\
& bi-directional LSTM & 7.38 & 30.77 \\
\hline \hline
Pronlex & past results~\cite{BisaniGraphone2008,Chen03G2P}  & 6.78 & 27.33 \\
& bi-directional LSTM & 6.51 & 26.69 \\
\hline \hline
\end{tabular}
\caption{The PERs and WERs using bi-directional LSTM in comparison to the previous best performances in the literature. }
\label{tab:final}
\end{table}


\section{Related Work}
\label{sec:related}
Grapheme-to-phoneme has important applications in
text-to-speech and speech recognition. It has been well studied in the
past decades. Although many methods have been proposed in the past,
the best performance on the standard dataset so far was achieved using
a joint sequence model~\cite{BisaniGraphone2008} of grapheme-phoneme
joint multi-gram or \textit{graphone}, and a maximum entropy model~\cite{Chen03G2P}. 

To our best knowledge, our methods are the first single neural-network-based system
 that outperform the previous state-of-the-art
methods~\cite{BisaniGraphone2008,Chen03G2P} on these common datasets. 
It is possible to improve performances by combining multiple systems and methods\cite{WuLTS2014,RaoSLTMG2P}, we have chosen not to engage in building hybrid models. 

Our work can be cast in the general sequence to sequence translation category, which includes tasks such as machine translation and speech recognition. Therefore, perhaps the most closely related work
 is \cite{SundermeyerTranslation}. However, instead of the marginal gains in their bi-direction models, our model obtained significant gains from using bi-direction information. Also, their work doesn't include experimenting with deeper structures, which we found beneficial. We plan to conduct machine translation tasks to compare our models and theirs. 




\section{Conclusion}
\label{sec:conclusions}
In this paper, we have applied both encoder-decoder neural networks and 
alignment based models to the grapheme-to-phoneme task. The encoder-decoder
models have the significant advantage of not requiring a separate alignment
step. Performance with these models comes close to the best previous 
alignment-based results. When we go further, and 
inform a bi-directional neural network models with 
alignment information, we are able to make significant advances over
previous methods.

\bibliographystyle{ieeebib}
\bibliography{refs,RNNLU}

\end{document}